\documentclass[conference]{IEEEtran}

\usepackage{times}
\usepackage{epsfig}
\usepackage{graphicx}
\usepackage{amsmath}
\usepackage{amssymb}
\usepackage{acronym}
\usepackage{microtype}
\usepackage{subfig}
\usepackage{tabularx} 

\usepackage{hyperref}

\acrodef{OCR}{Optical Character Recognition}
\acrodef{HMM}{Hidden Markov Model}
\acrodef{OAA}{One Against All}
\acrodef{OAO}{One Against One}
\acrodef{kNN}{k-Nearest Neighbour}
\acrodef{NN}{Neural Network}
\acrodef{SVM}{Support Vector Machine}
\acrodef{NB}{Naive Bayes}
\acrodef{PW}{Parzen Window}
\acrodef{MVMN}{Multivariate Multinomial Distribution}
\begin{document}

%%%%%%%%% TITLE
\title{Augmentation of base classifier performance via HMMs on a handwritten character data set}

\author{
Helder F. S. Campos and Nuno M.C. Paulino\\
{\tt\small helder.campos@fe.up.pt, nmcp88@gmail.com}\\
DEEC, FEUP\\
Rua Dr. Roberto Frias, s/n 4200-465 Porto PORTUGAL\\
}

\maketitle
\thispagestyle{empty}

%%%%%%%%%
\begin{abstract}

This paper presents results of a study of the performance of several base classifiers for recognition of handwritten characters of the modern Latin alphabet. Base classification performance is further enhanced by utilizing Viterbi error correction by determining the Viterbi sequence. \acp{HMM} models exploit relationships between letters within a word to determine the most likely sequence of characters. Four base classifiers are studied along with eight feature sets extracted from the handwritten dataset. The best classification performance after correction was 89.8\%, and the average was 68.1\%.

\end{abstract}

%%%%%%%%%
\section{Introduction}

\ac{OCR} is the process of identifying handwritten or typewritten text, including either characters or digits, in order to automatically translate this input into a digital storing format. The application scenarios of this concept are vast, involving any field where digitalization of documents for the purpose of facilitating data storage and finding documents applies.

OCR is thus a problem within the domain of machine learning, wherein a set of input data (images) requires classification into different classes (words or characters). Since an efficient conversion of documents from text to digital format is desired, robust classifying mechanisms must be developed. Because handwritten text is highly variable (depending on the author) and captured images of text may contain noise such as smudges or pen strokes which are accidental or incomprehensible, constructing a good classifier requires a solid understanding of the characteristics of the data. Complex character sets, such East Asian alphabets, are an example of the difficulty, due to the similarity and number of characters they contain.

OCR may function online or offline. Offline recognition is the conversion of static, already printed/written documents. Online OCR is the detection of real\-time writing, which greatly increases the required response time and, consequently, lowers the allowed computation complexity which in turn lowers performance.

This document reports the results of a short study performed on a handwritten dataset of characters, detailing extracted features and presenting classification accuracy for a set of base classifiers. The base classifier performance is then enhanced by utilizing \acp{HMM}, which exploit relationships between neighboring letters within words. 
This paper is organized as follows: Section~\ref{sec:stateart} briefly presents related works, Section~\ref{sec:overview} outlines the methodology, Section~\ref{sec:dataset} presents the utilized data set, Section~\ref{sec:features} discusses feature extraction/selection performed on the data set, Section~\ref{sec:classifiers} characterizes the utilized classifiers, Section~\ref{sec:markmodel} explains the utilized \ac{HMM} models, Section~\ref{sec:results} presents the classification results and Section~\ref{sec:conclusions} concludes the paper.

%%%%%%%%%%%%%%%%%
\section{Related Work}\label{sec:stateart}

%Handwritting recognition can be oriented either towards recognizing the writting style of a single writter, or to be as generic as possible in order to classify any handwritten input, depending on the application.

A comprehensive study of handwriting recognition on 4 distinct alphabets (numerals, pairs of numerals and English capital letters and Korean characters) is presented in~\cite{suen}. An approach utilizing multiple \acp{NN} for binary decisions is utilized with about 250 input features, and is compared to the performance of a single \ac{NN}. A recognition rate of 91.11\% for the binary networks is achieved versus 81.03\% for the test set of the English capital letters. For Korean characters, these values are 68.75\% and 22.46\%.

%Mohand et al.~\cite{Mohand} propose a system in which online HMM adaptation is augmented by not only adapting the the HMM structure (number of states) but also the 

In~\cite{Velagapudi}, a very similar study to the one presented in this document is performed. The same data set is used and 10-fold cross validation techniques are employed for the implemented \ac{kNN} classifier, 2 layer \ac{NN} (50 hidden and 26 output nodes) and \ac{SVM}. The work the author presents differs from the one here presented in the \ac{HMM} model employed. The \ac{HMM} is built by measuring the number of errors that the base classifier makes while classifying each letter and feeding that information to the \ac{HMM} in the form of emission probabilities. In the end, the emission probabilities will be the probability of choosing letter $i$ knowing the true letter $j$, being the observations the classified letter. This approach, being simple, has the disadvantage of losing information of the exact shape of the letter classified. Because of this, our approach tries to capture the uncertainty that the classifier has, not in a general letter class, but in a specific letter. How this is done will be explained latter. Nevertheless, an average of 5,5\% classification improvement is reported with \ac{HMM} error correction, and an average 69\% base classification recognition rate, the maximum being 88\% using a \ac{SVM}.

%%%%%%%%%%%%%%%%%
\section{Overview}\label{sec:overview}

Typical pattern recognition approaches involve: pre-processing, feature extraction/selection, training a classifier and testing the classifier. 

We performed limited pre-processing on the data set. Features, discussed in section~\ref{sec:features}, were extracted from the original dataset and then we determined their potential usefulness analyzing whether or not they were distinctive from class to class. We trained several base classifiers with several feature combinations, and then utilized the probabilities calculated by these classifiers as emission probabilities for the \ac{HMM} models. We aimed to analyze the performance of each base classifier, and which one benefited most out of \ac{HMM} error correction.

%%%%%%%%%%%%%%%%%
\section{OCR Data Set}
\label{sec:dataset}

The dataset utilized contained 52152 handwritten letters, totalling 6877 words split into 10 validation folds. The same words are repeated throughout the dataset in different folds (and different handwriting), so the effective number of different words is 55. In average, each word is repeated 124 times in the data set, with the least occurring word having 70 instances and the most occurring 150 instances. Capital letters, digits or punctuation are not included in the set. Letters from \emph{a} to \emph{z} are included in the set. Each letter is normalized to retain relative size and represented by a 8x16 (128) black and white (binarized) pixel image. Each letter has an associated word \emph{ID}, and letters in sequence within the data set with the same word \emph{ID} compose a single word. The first (capital) letter of each word has been removed. The dataset was collected by Rob Kassel at MIT Spoken Language Systems Group~\cite{OCRDataSet}. 

\begin{figure}
  \centering
  \subfloat[A recognizable \emph{a}\label{fig:goodA}]{
    \includegraphics[width=0.20\linewidth]{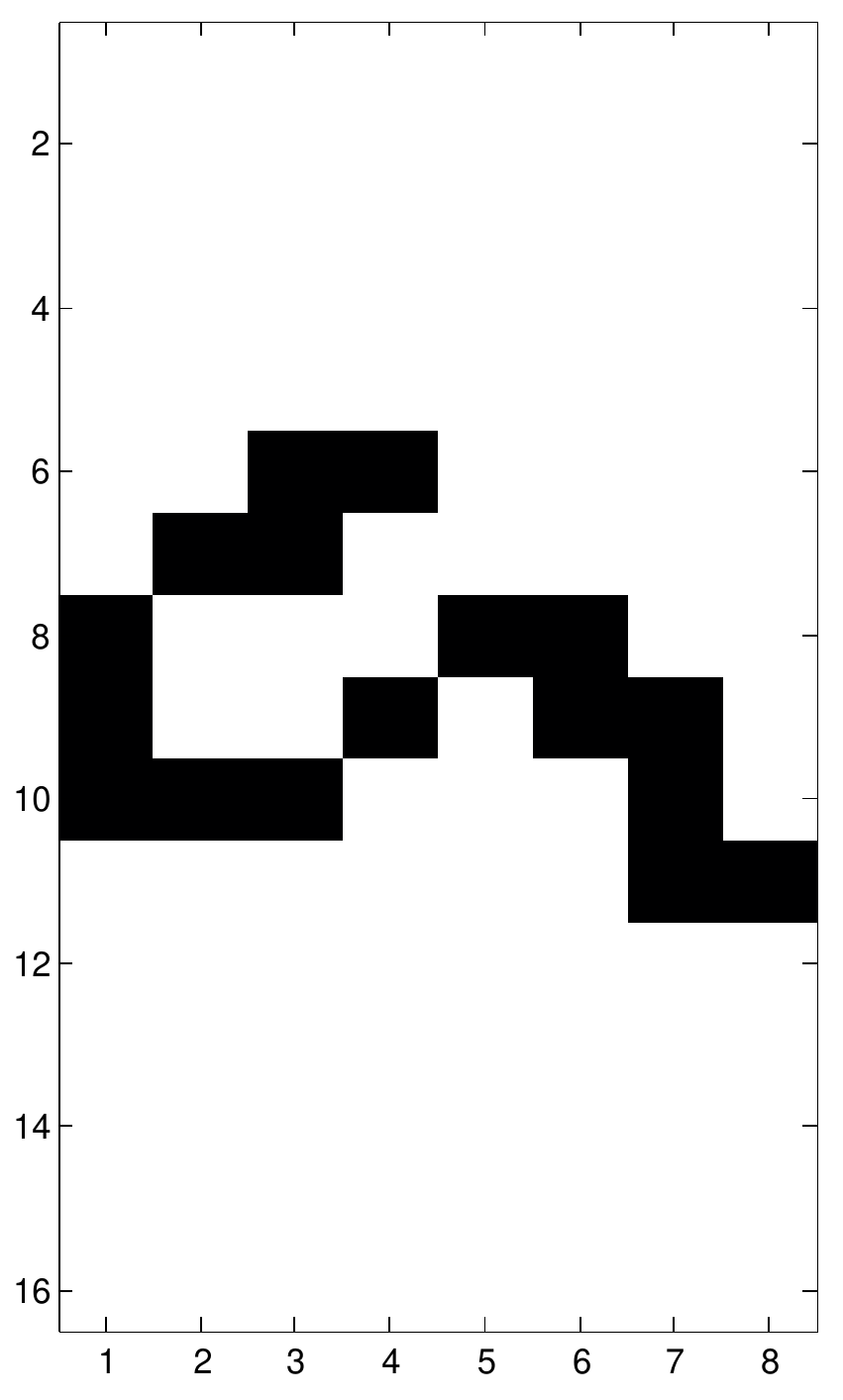}
  }
  \hspace{5mm}
  \subfloat[An unrecognizable \emph{a}\label{fig:badA}]{
    \includegraphics[width=0.20\linewidth]{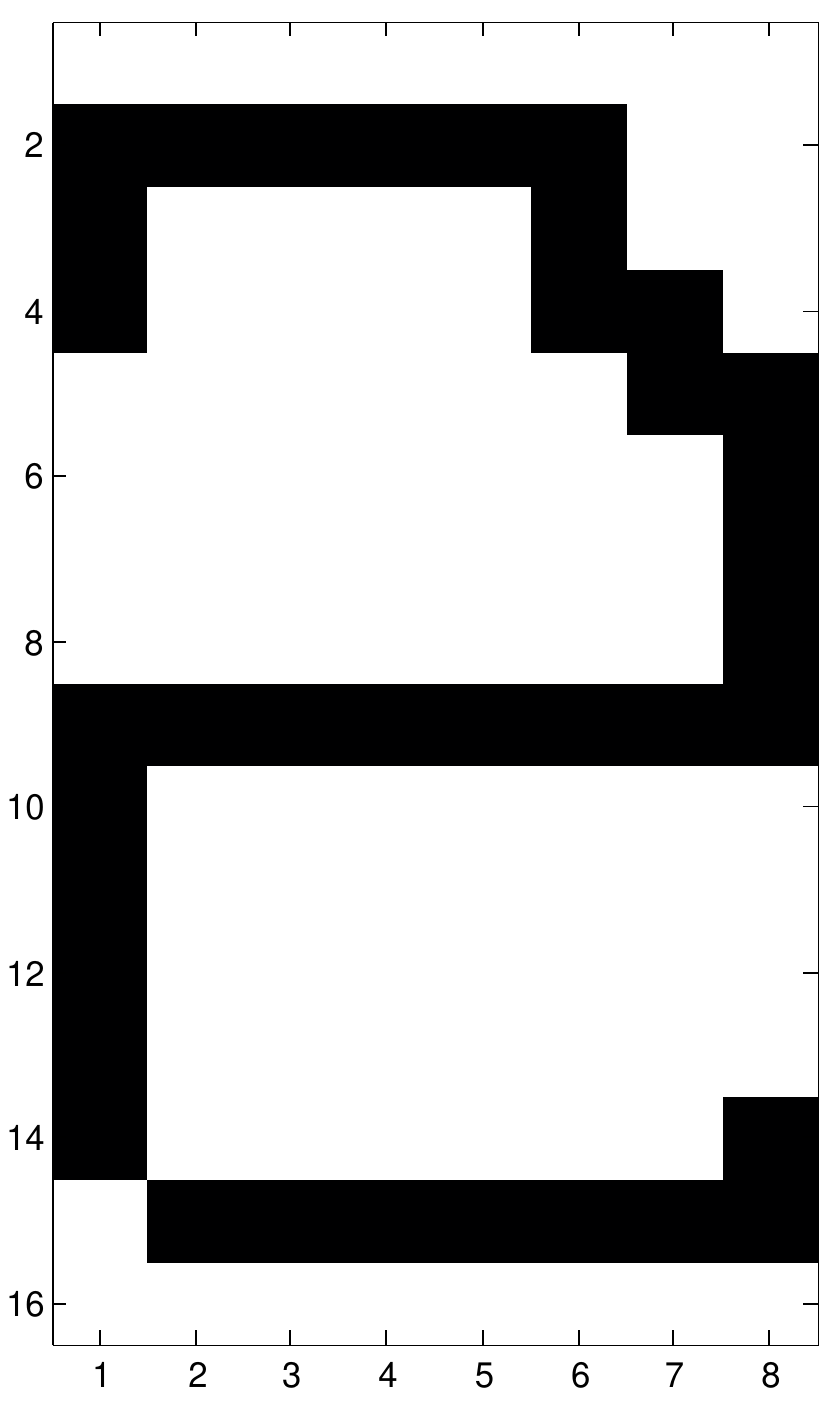}
  }
  \hspace{5mm}
  \subfloat[An outlier \emph{g}\label{fig:strangeG}]{
    \includegraphics[width=0.20\linewidth]{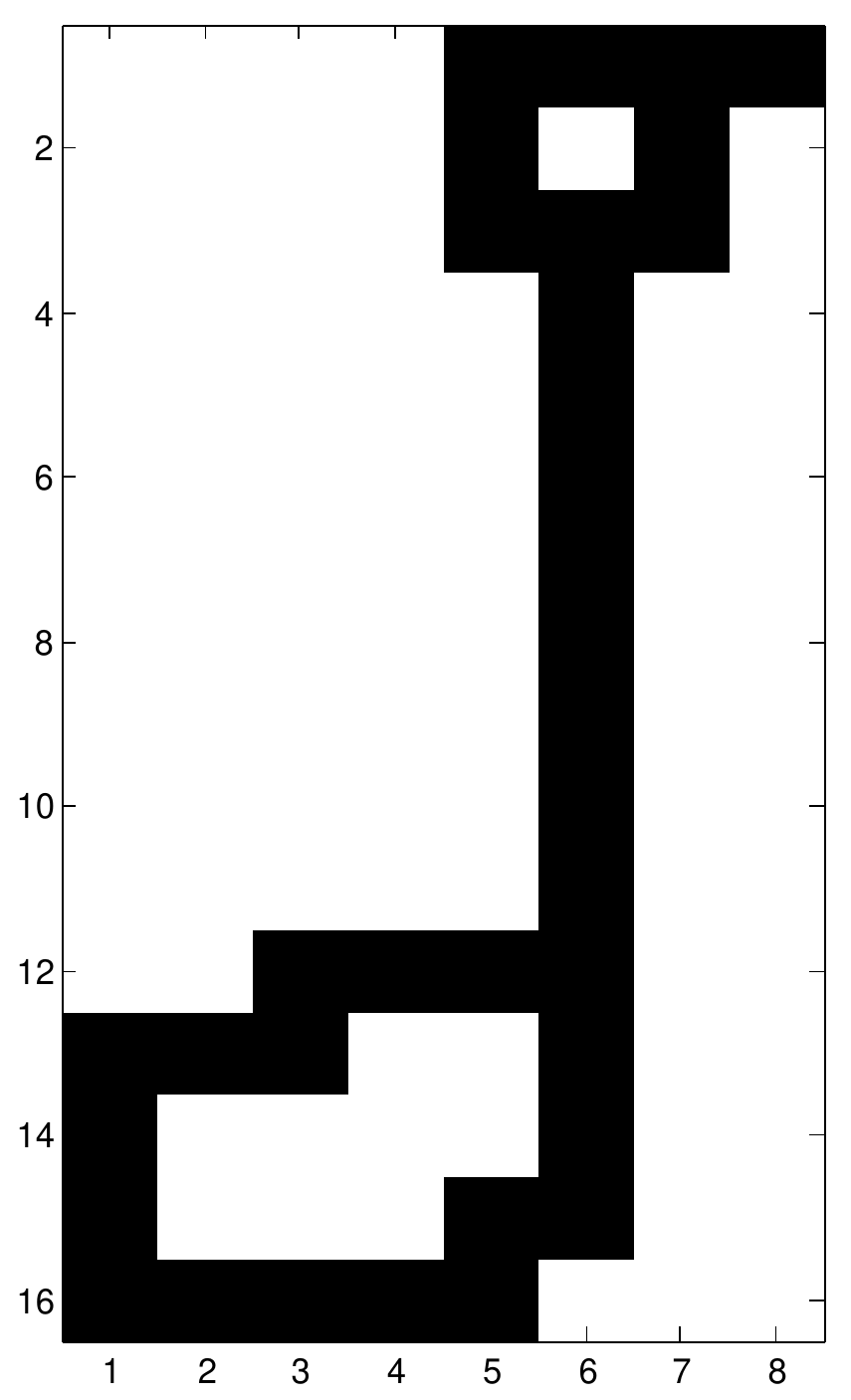}
  }  
  \caption{Three letters from the dataset}\label{fig:letters}
\end{figure}

Despite this preprocessing, there is considerable variation per letter within the data set. Figure~\ref{fig:letters} represents 3 letters extracted from the set. Figure~\ref{fig:goodA} is an example of a character easy to recognize, while figure~\ref{fig:badA} does not resemble the character it represents. Other characters, while recognizable by inspection, vary considerably from others in the same class within the data set, such as~\ref{fig:strangeG}, which is reasonably different from other \emph{g}s. Even the \emph{a} in figure~\ref{fig:goodA} has this behaviour in the sense that it is not \emph{"closed"}, whereas many \emph{a}s have a hole.

Although the dataset was divided in a typical value of 10 cross-validation folds, the utilized neural network toolbox implemented a 1 fold hold-out method, i.e., it divides the data into three sets: training, validation and test. With this method, the training time for 26 binary networks (which we choose as our \ac{NN} model for reasons explained in section~\ref{networks}) during some preliminary evaluations took around 40 minutes. If were were to utilize the 10 folds for cross-validation the required time would be large. So, to implement the networks in reasonable time, we utilized the default toolbox method of 1 hold out validation set. For validity of comparison, we partitioned the data in the same manner for the rest of the classifiers. The train, validation and test sets are the same for all combinations of classifier and input features. In order to ensure that the data is evenly split (i.e., all three sets contain evenly split samples of all the letters), we developed a mechanism that randomly selected the available folds to create the three datagroups, by specifying the division ratios. For ratios that resulted in partial folds being required, we would instead scramble and distribute the remainder folds for the three groups. The scramble was made at word level, preserving the letter sequence in the data. We utilized ratios of 1/3 for each set, which resulted in train, validation and test sets of 17214, 17548, 17376 samples respectively.

% If folds could not be made divided using multiples of the 10 subsets we would scramble the not fully used subset and divide it between two new sets. 

% Whenever the division could not be made by using multiples of the 10 subsets we would scramble the not fully used subset and divide it between two new sets, keeping in mind that data (letters) ha
 
 % we used the already made division of the data to create the sets, randomly selecting those already divided subsets. 

% After this data division, the developed framework proceeded to test specified feature/classifier combinations, and saved the resulting train and test accuracies, along with train and test times. The resulting probability distributions are then usd

%%%%%%%%%%%%%%%%%
\subsection{Preprocessing}

Pre\-processing the inputs of a classifier is done to try and reduce the variability of a feature that is known not to contain useful information or that, in turn, distorts useful information.

\begin{figure}
  \centering
  \subfloat[]{
    \includegraphics[width=0.20\linewidth]{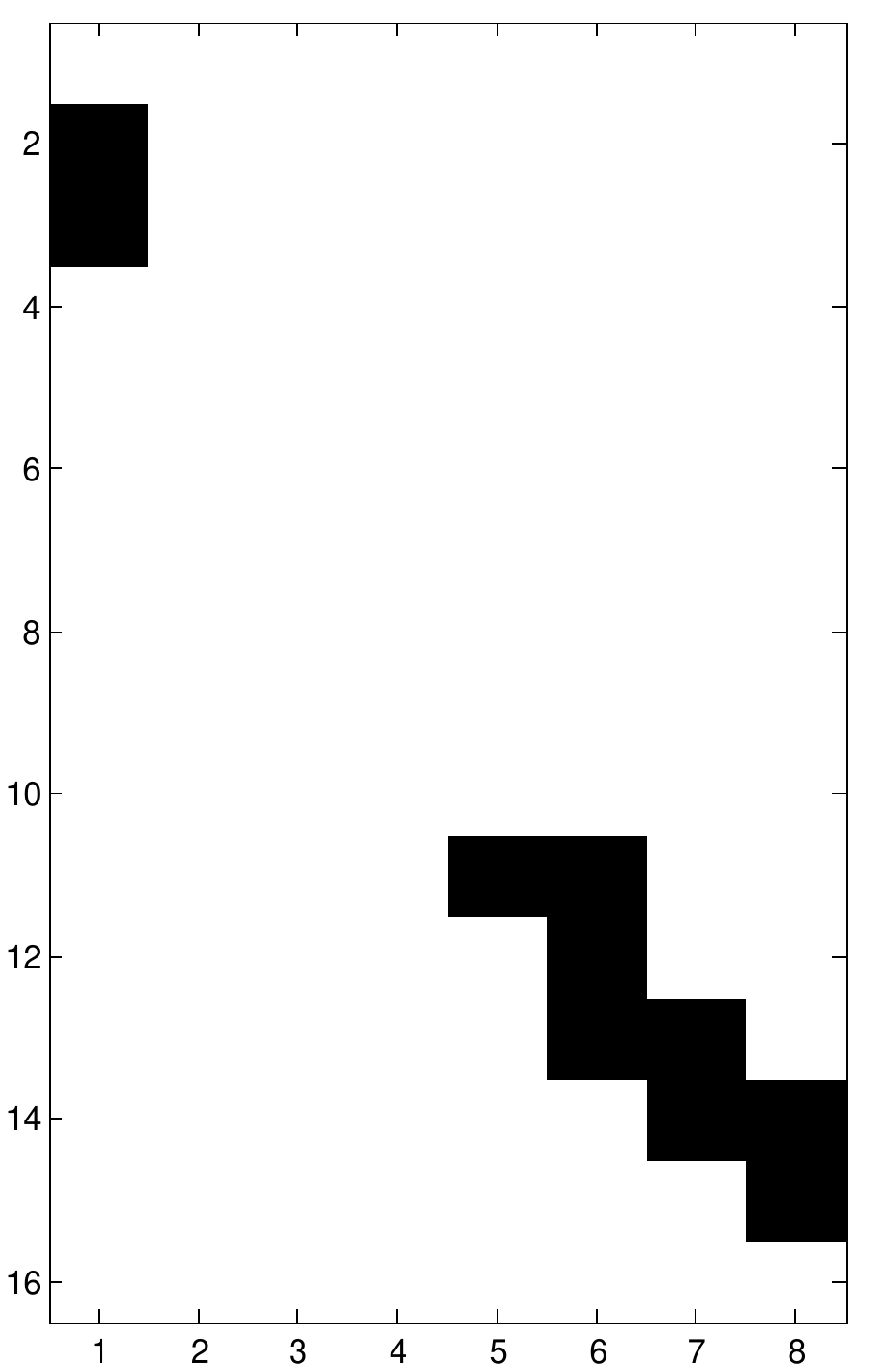}
  }  
  \hspace{1mm}
  \subfloat[]{
    \includegraphics[width=0.20\linewidth]{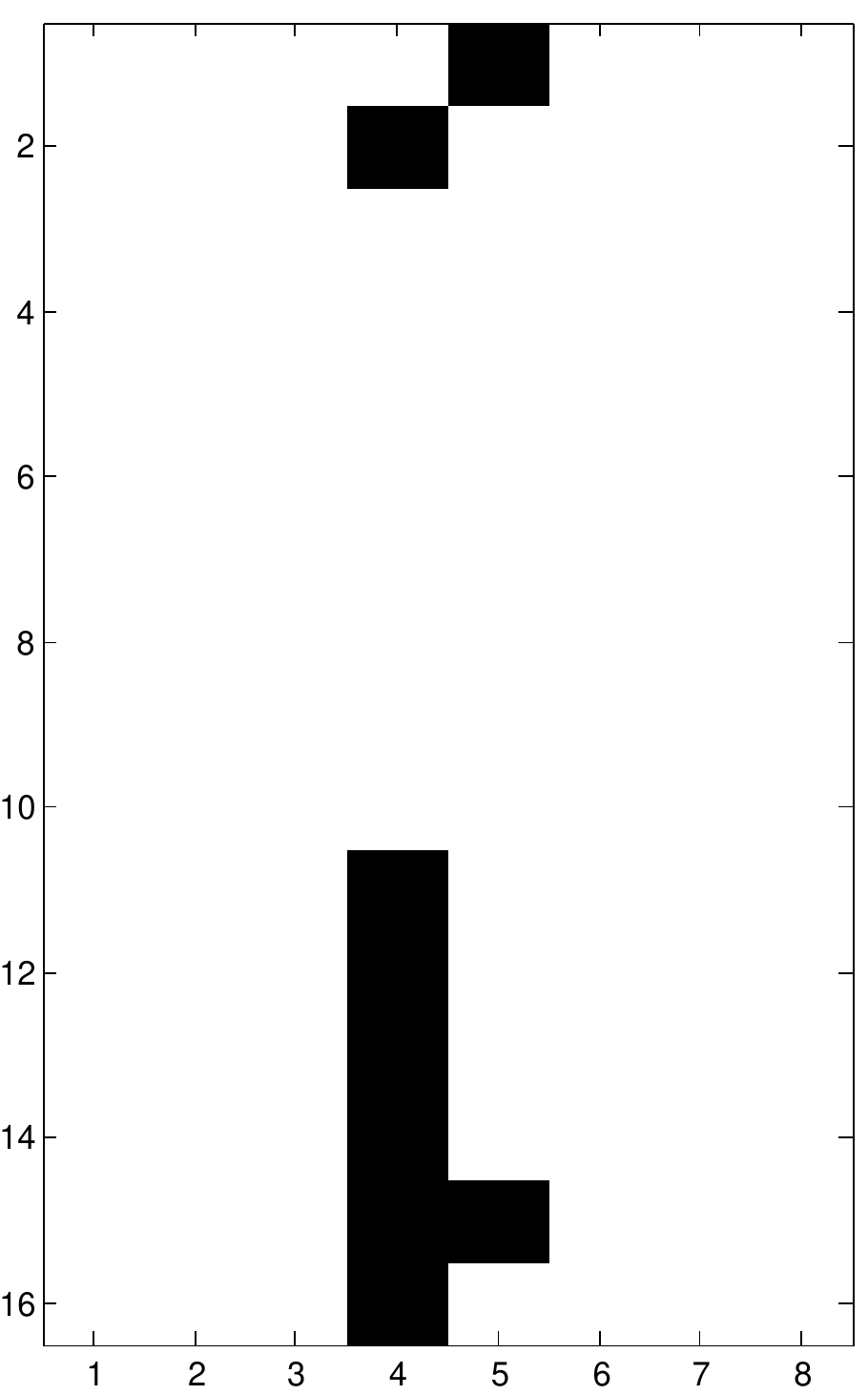}
  }  
  \hspace{1mm}
  \subfloat[]{
    \includegraphics[width=0.20\linewidth]{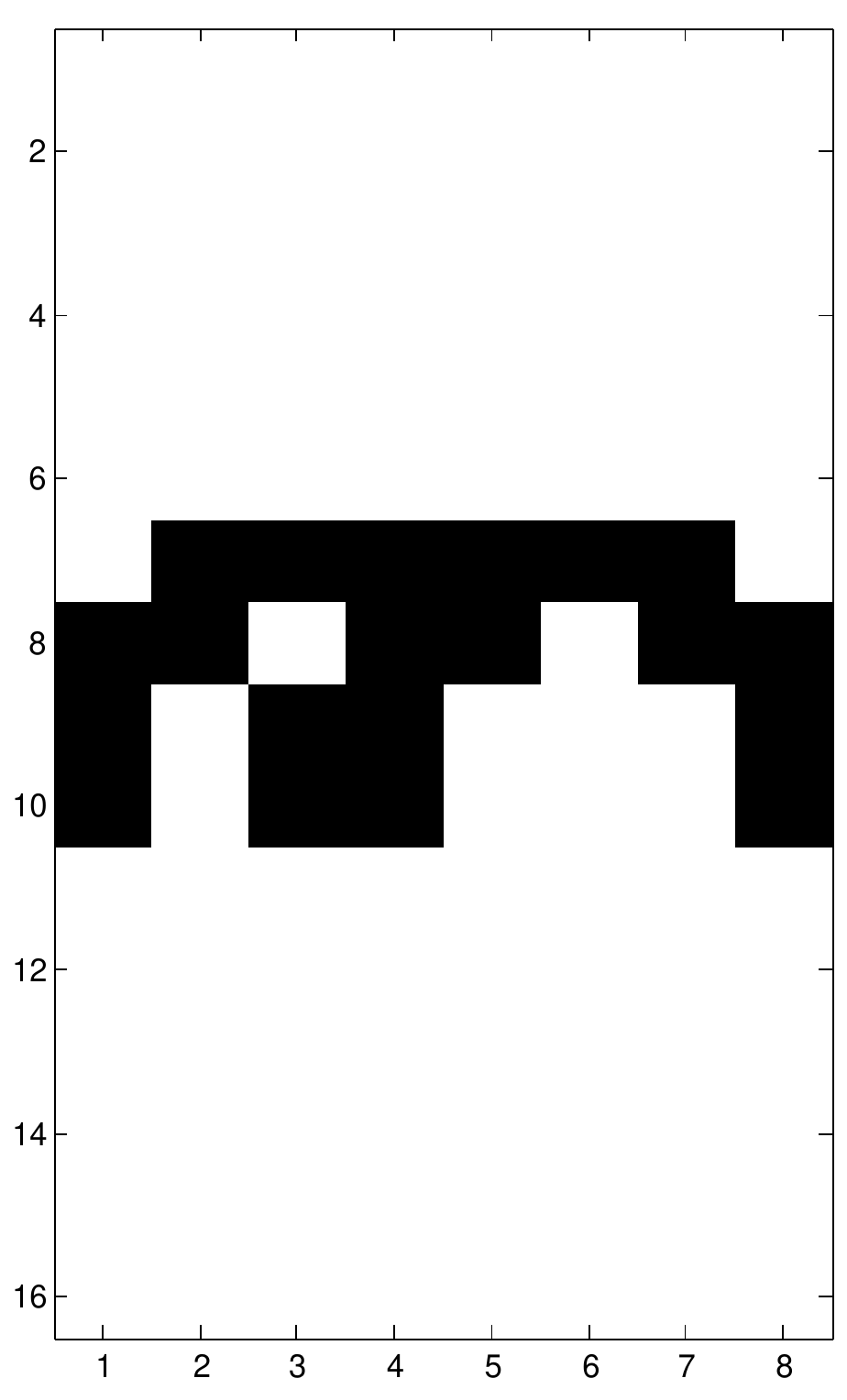}
  }
  \hspace{1mm}
  \subfloat[]{
    \includegraphics[width=0.20\linewidth]{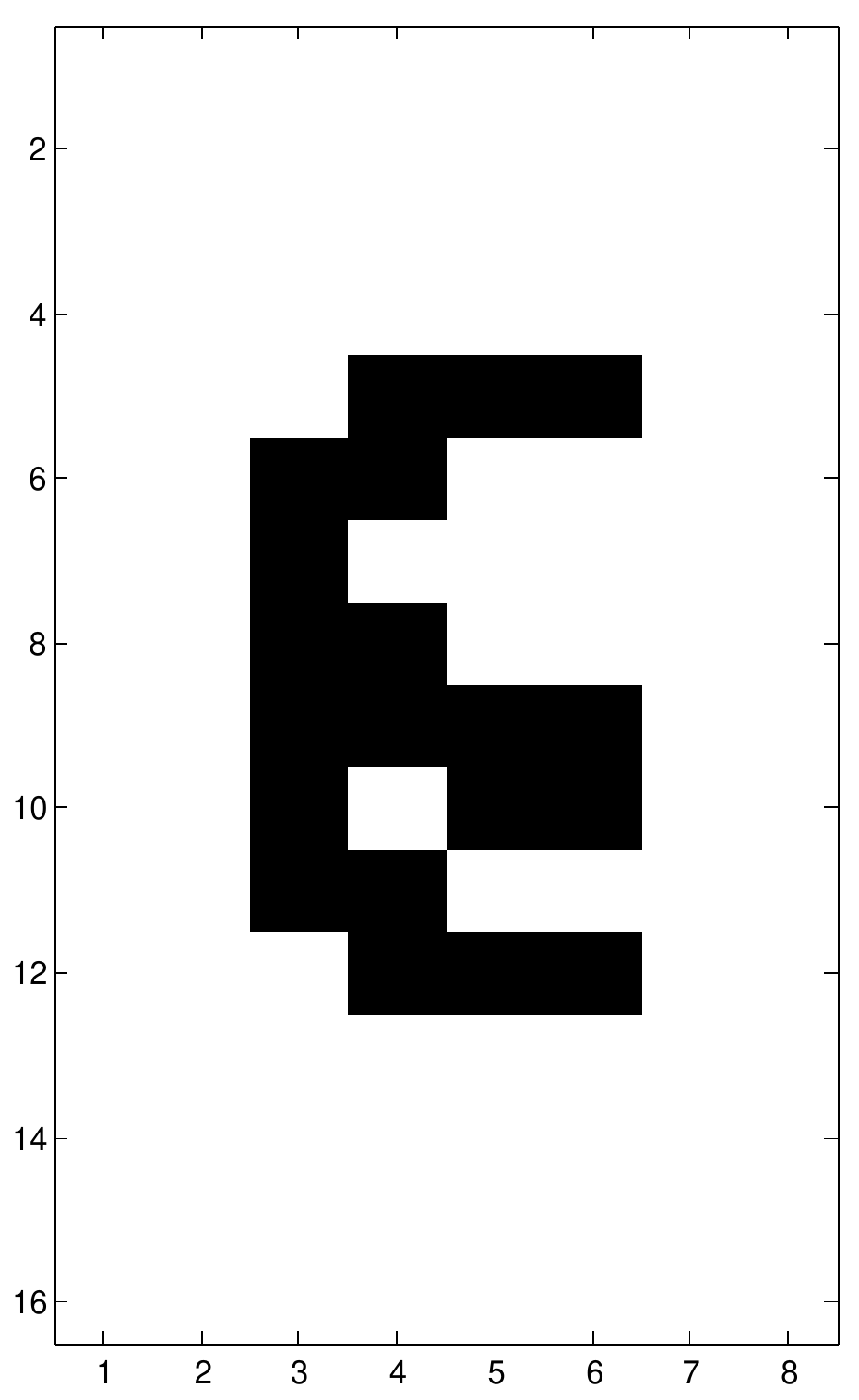}
  }
  \caption{Preprocessing of the input pixel maps}\label{fig:unslant}
\end{figure}

After verifying that many letters were slanted, we attempted rotating the letters in order to obey the natural orientations of the words (\emph{l}s and \emph{i}s are vertical while \emph{m}s are horizontal). However, to better orthogonalize the inputs to the classifiers, orientation information was removed from the pixel maps altogether. It was given to the classifiers as a separate input. As explained in the following section, the orientations we could extract were those of the longest axis of the letter. We rotated this axis upwards, effectively removing the information from the pixelmap. Figure~\ref{fig:unslant} exemplifies this procedure, 

Note that rotation sometimes resulted in loss of pixels, due to low resolution. Scaling the image before rotating and then scaling the result down does not improve this behaviour and in fact introduces fictitious information by scaling up the image. So, because the dataset has already been considerably preprocessed, any further modification would potentially reduce the information. One possible alteration would be the removal of outlier pixels (some letters have noise pixels or have superfluous strokes). This would involve robust outlier detection which falls out of the scope of this study.

%%%%%%%%%%%%%%%%%
\section{Feature Extraction}\label{sec:features}

\subsection{Pixels as features}

Some of the classifiers were trained directly with the 128 pixels representing the letter. Despite being a much larger feature space than the extracted features explained further, it is still a much smaller number of inputs when compared to some approaches that use a feature vector of 200 or more elements~\cite{suen}.

\subsection{Extracted features}

\begin{figure*}[!h]
  \centering
  \subfloat[]{
    \includegraphics[width=0.47\linewidth]{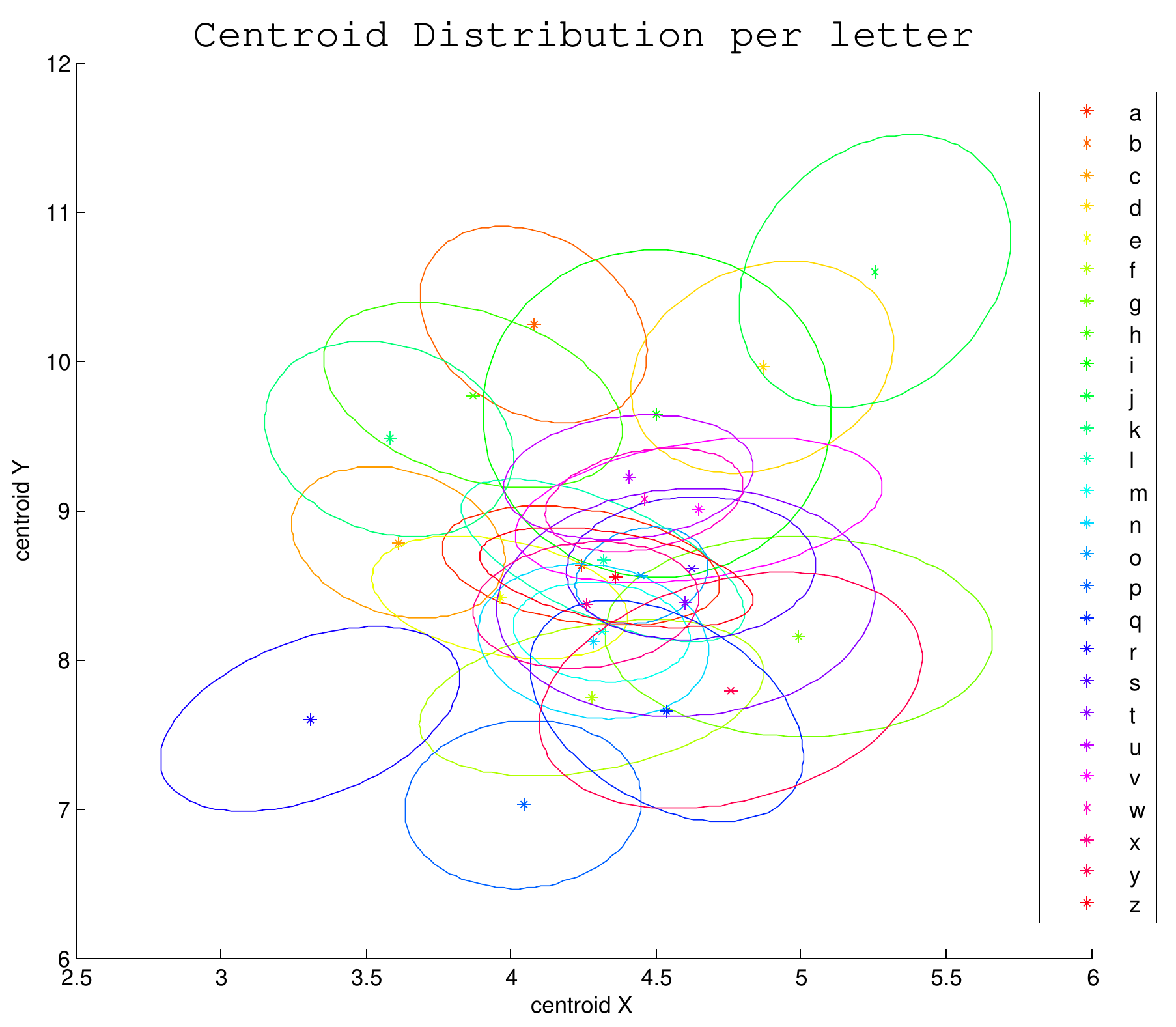}
    \label{fig:centroid}
  }
 % \hspace{5mm}
  \subfloat[]{
    \includegraphics[width=0.47\linewidth]{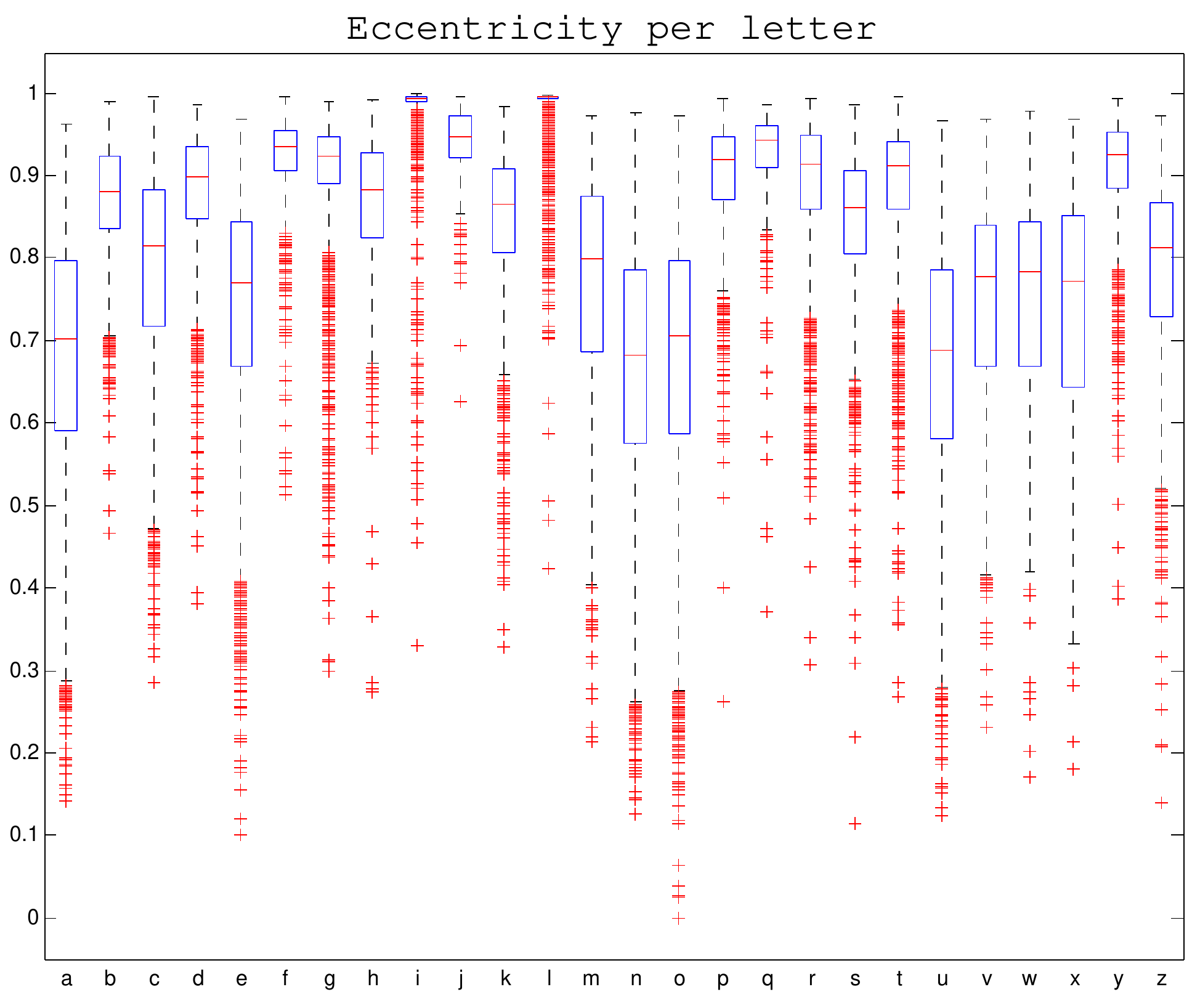}
    \label{fig:eccent}
  }
  
    \centering
  %\hspace{5mm}
  \subfloat[]{
    \includegraphics[width=0.47\linewidth]{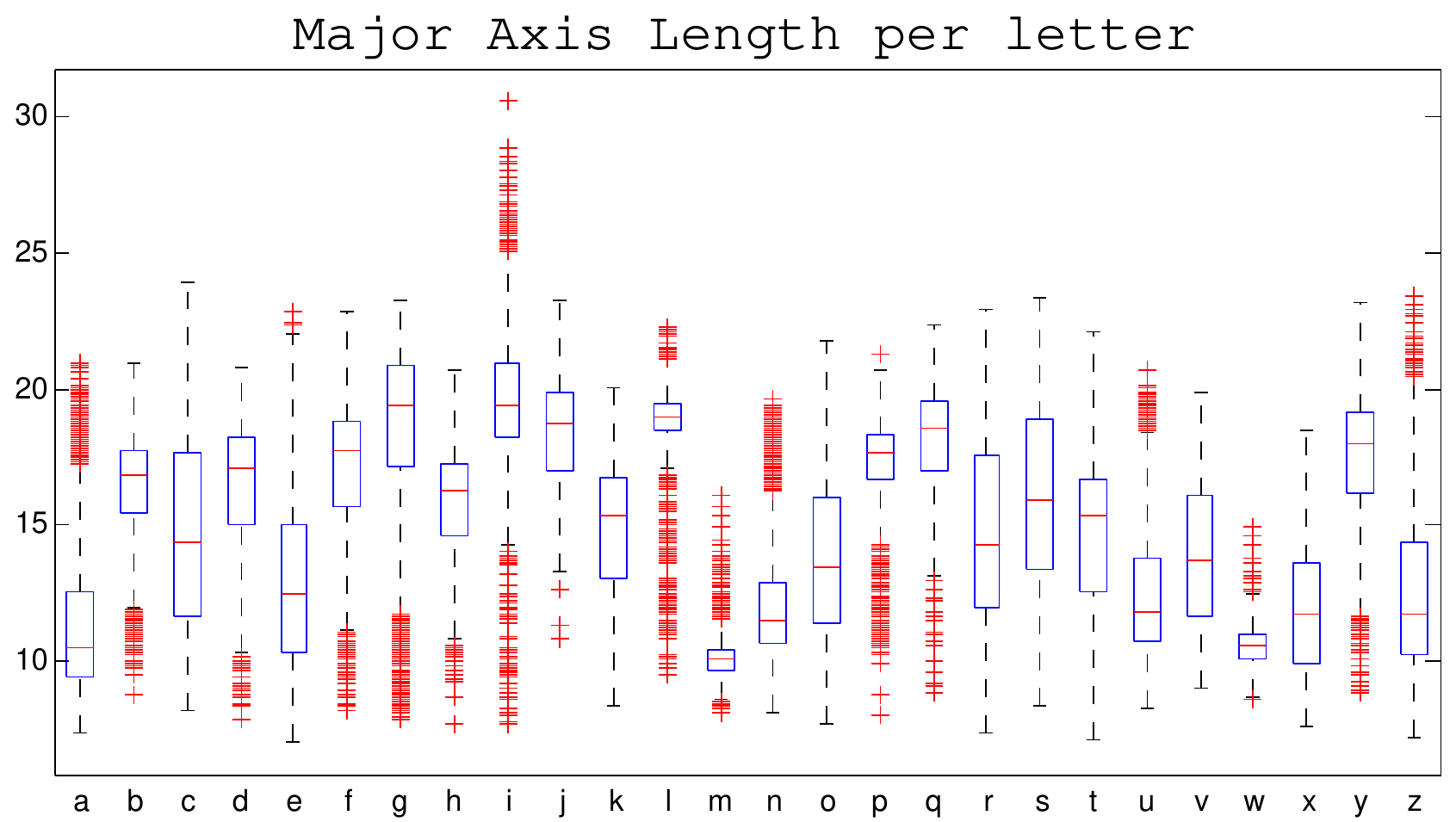}
    \label{fig:majorax}
  }  
  %\hspace{5mm}
  \subfloat[]{
    \includegraphics[width=0.47\linewidth]{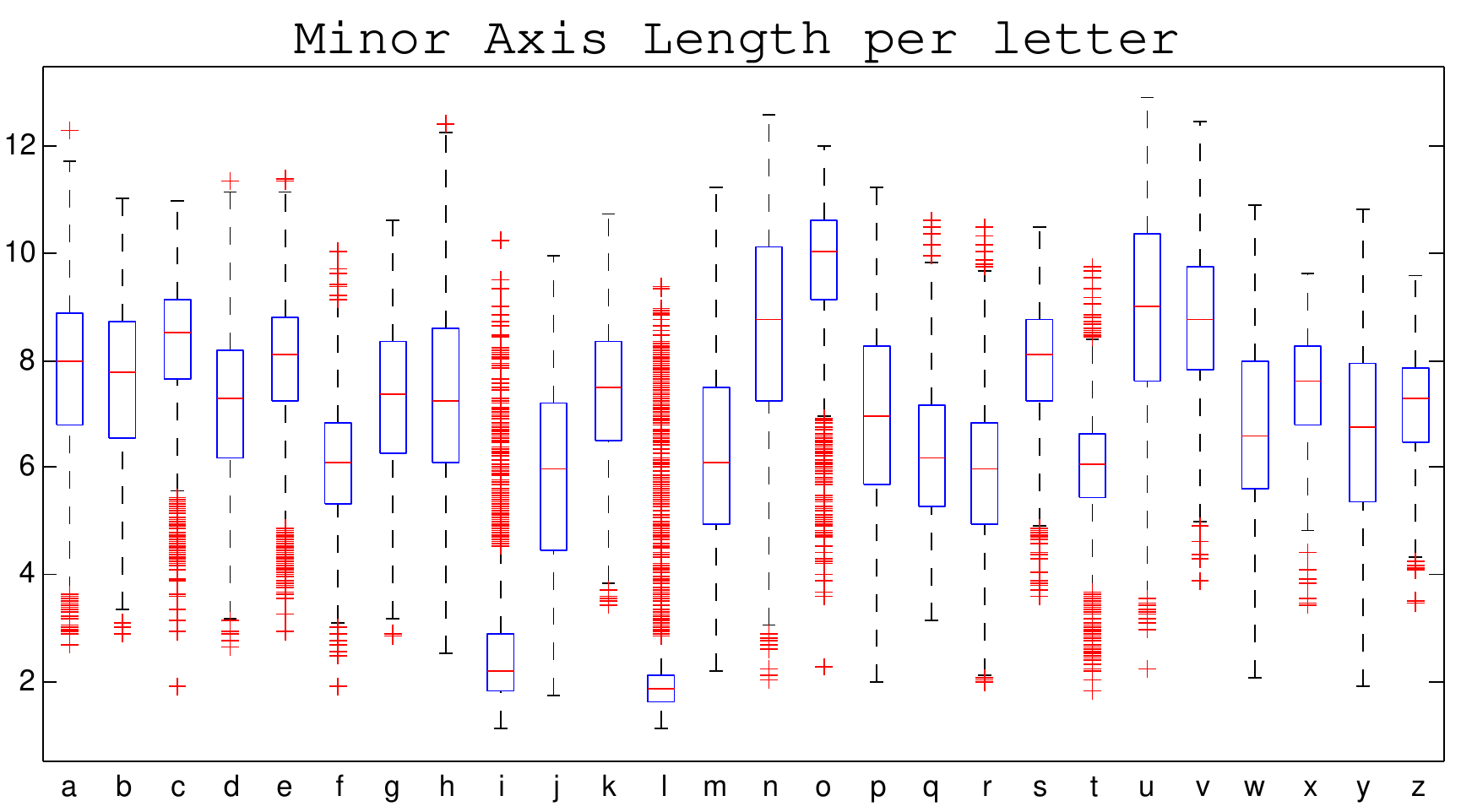}
    \label{fig:minorax}
  }  
  
    \centering
  %\hspace{5mm}
  \subfloat[]{
    \includegraphics[width=0.47\linewidth]{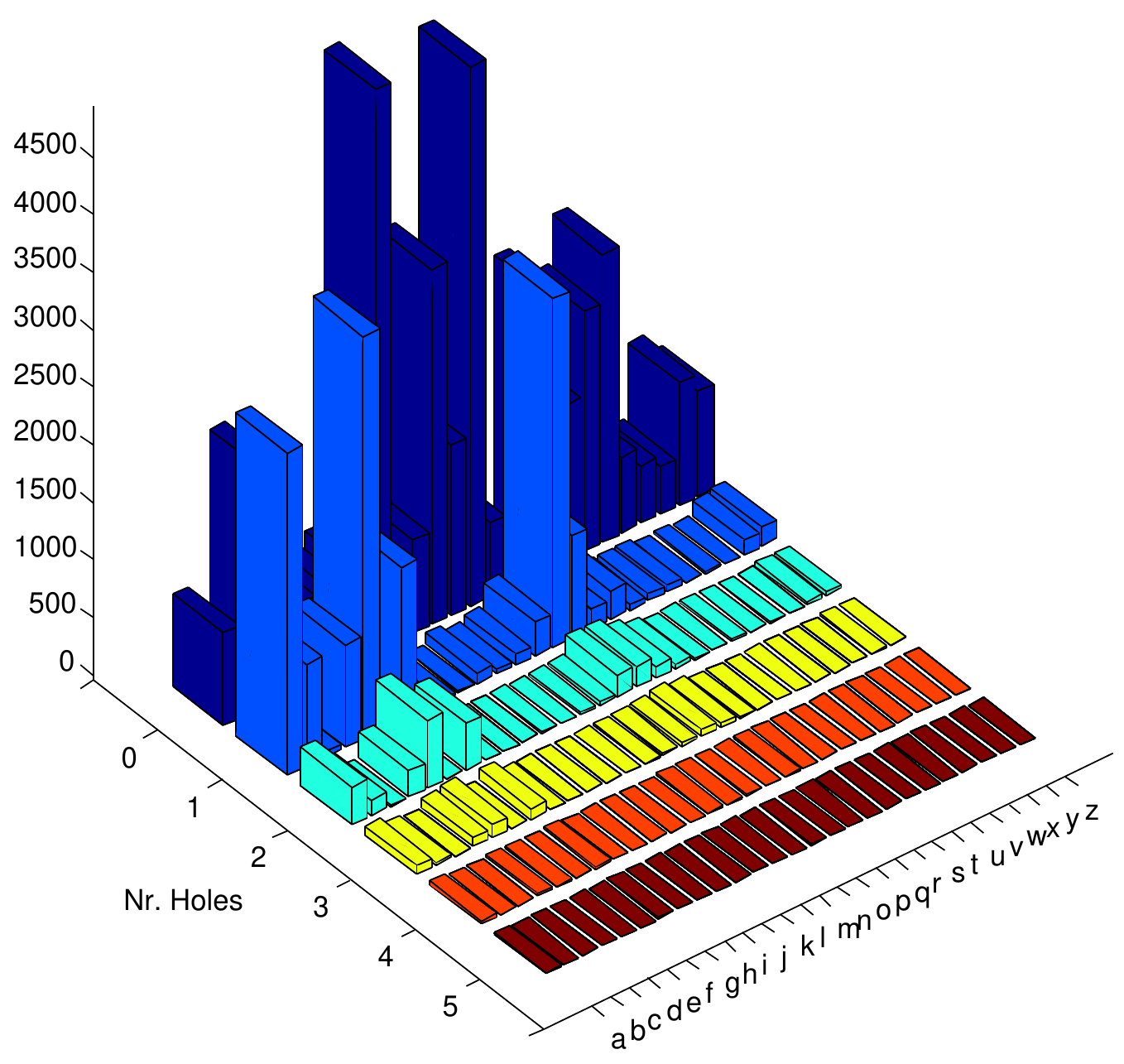}
    \label{fig:nrholes}
  }  
  %\hspace{5mm}
  \subfloat[]{
    \includegraphics[width=0.47\linewidth]{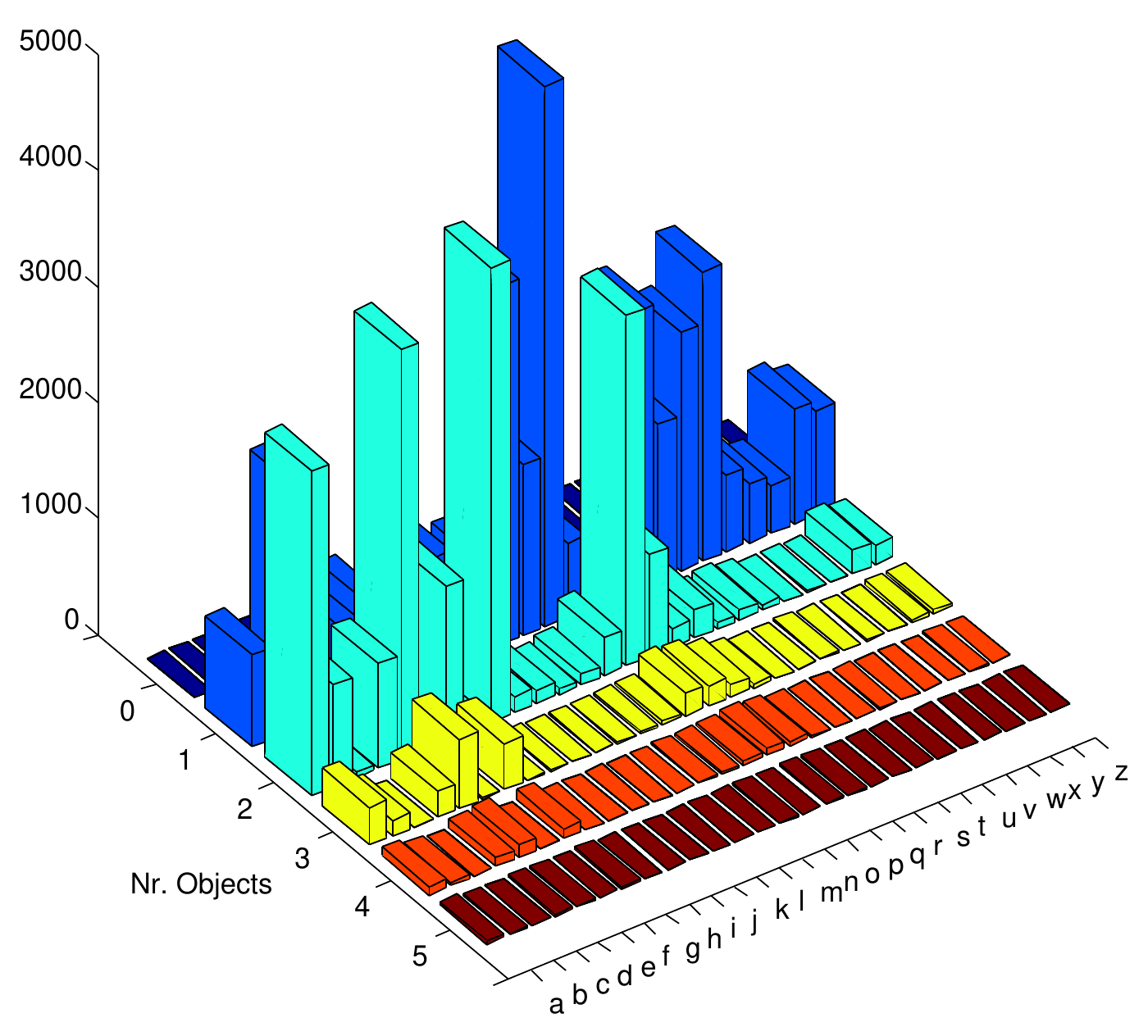}
    \label{fig:nrobjs}
  }  
  \caption{Features of the dataset}\label{fig:features}
\end{figure*}

Studying custom feature extraction fell out of the scope of this paper, so, several features were extracted from the pixel maps utilizing a MatLab region properties recognition toolbox. The features analyzed were: the orientation of the ellipse encompassing the letter, the length of its major and minor axis and its eccentricity as well as its centroid; the number of pixels in each subquadrant of 4x8 pixels in the character; the number of objects and holes in the letter; several metrics involving the convex hull of the letter and its area as well as the area occupied by active pixels and the perimeter of the letter. Some of these features are represented in figure~\ref{fig:features}, discriminated per letter. These are features as extracted from the original unprocessed data set.

Normalization performed on the images allowed for a common reference center to be established for all letters, thus making features such as the centroid, show in figure~\ref{fig:centroid}, more distinctive. The mean of the centroid of the ellipse that characterizes each class is shown, with each axis representing pixels. The lines shown are one standard deviation. Although there is a great overlap of distributions for many letters, some are distinctly different, which makes this a characterizing feature. Figure~\ref{fig:eccent} depicts the boxplots of the eccentricity of the ellipse for each letter. This is a measure of the letter's shape. Greater eccentricity (maximum of one, meaning a line segment) means a thinner letter, while low eccentricity (minimum of zero, meaning a circle) represents wider letter. Since the data set is scaled and, by nature, many letters have similar bounding boxes, this does not help distinguish every individual letter. However, it is discriminant enough to intuitively separate letters into either thin or wide. For instance, from \emph{p} to \emph{t}, the distribution of eccentricities is very distinctive from the neighboring letters. Distributions of the values for the major and minor axes of the ellipses for each letter is shown in Figures~\ref{fig:majorax} and~\ref{fig:minorax}. Since eccentricity is directly related to the relation of the two values, their boxplots are similar to that of the eccentricity. That is, they cannot directly discriminate each letter but some have distinctive values.

The number of holes and objects for each letter is show in Figure~\ref{fig:nrholes} and~\ref{fig:nrobjs} respectively. The observed numbers have some deviation from what would be expected. Due to handwriting and low resolution, a thinly written \emph{a}, for instance, might separately isolate 3 pixels within its bounds, reporting 3 holes. In the same way, the number of objects may be misreported if pen strokes are not in contact.

Some features were not considered appropriate as classifier inputs. For instance, the bounding box was observed to be far to similar amongst all letters, partially due to the fact that they are normalized and also because a bounding box would represent a \emph{g}, for instance, as square and a \emph{t} as well. Any other measurement involving area, such as the area of the convex hull or the total number of pixels, was also too similar between classes. To make the number of active pixels more useful, we split the region into 4 subregions, as stated, which results in a more distinctive count of pixels per quadrant, as the shape of each letter influences the location of the pixels.

A total of eight feature combinations were fed to different classifiers to determine performances and training times for each scenario. These are detailed along with the results in Table~\ref{tab:resultstab}.

\section{Hidden Markov Model}\label{sec:markmodel}

An \ac{HMM} is a probabilistic model that represents sequences of observations by considering that the probability of each single observation is influenced by the state of the system. The state is an unobserved (hidden) variable that alters the probability of the observation itself, such that a particular sequence of observations may have been generated by one possible corresponding sequence of associated states. In this context, observation probabilities are known as emission probabilities (i.e., the probability of a given state emitting an observation). If we wish to compute the probability of observing a certain event or feature after a previous sequence (e.g.,  the probability of drawing a particular card from a set after drawing others before), we need to compute all the possible sequences of states that could have generated that sequence. This would generate a set of priors, each indicating the probability of currently being in a particular state. These probabilities, in turn, define probability of observing that sequence regardless of the hidden states.

However, in the case of character detection, we are interested only in the sequence that was most likely to generate our observations. This is word level modeling of the character sequence, although \acp{HMM} can model individual characters.

So, we desire the sequence of hidden states (characters) most likely to have generated observed features (pixels or others). This is known as the Viterbi path, and improving classification via this method is known as Viterbi error correction. The base classifier outputs are our emission probabilities and with \ac{HMM} error correction, classification ambiguity can be solved with the knowledge that an \emph{l} does not transition to a \emph{c} for instance, but will do so for an \emph{o}. In this case, classification error would occur due to the similarities between the letters \emph{c} and \emph{o}.

In order to model these sequences the probabilities of transitioning from one state to another are needed. For \acp{HMM} these values are held by a transition matrix $A$, in which each element $A_{i,j}$ denotes the probability of changing from state \emph{i} to \emph{j}. To compute this matrix we utilized the train partition of the data set and counted the number of times each letter transitioned to another. Logically, only transitions between letters in the same word were considered. This results in a 26 by 26 matrix that holds these counting values, that we then normalized row wise by the sum of each same row. The following are the first 3 rows and 4 columns of the transition matrix (where values are subject to the randomly divided data set). 

%show some values of the transition matrix, proving the 0 transition probabiliy in some cases

\[ \left( \begin{array}{ccccc}
  & a & b 	& c 		& d\\
a & 0 & 0.1370 	& 0 		& 0.0288\\
b & 0 & 0 	& 0.1855 	& 0.0890\\
c & 0 & 0 	& 0.0981 	& 0\end{array} \right)\] 

Values at 0 indicate that some transitions do not occur, which effectively reduces the computational complexity of the \ac{HMM} and models the word level behaviour. A problem that might arise with the computation of this matrix is that if the data set does not represent a wide thesaurus, some transitions will never be modeled. In this matrix, $A_{a,c}$ was computed as 0 while this transition does in fact exist in real words.

In order to fully define the \ac{HMM} model, the initial state probabilities are required. These represent the priori probabilities of the first letter being a given one in the alphabet. Similarly, these were computed empirically by counting the number of occurrences of each letter at the beginning of a word and normalizing by the count. These probabilities, the matrix $A$ and the emission probabilities define the \ac{HMM} model.

\begin{figure}[!t]
\centering	
	\includegraphics[width=0.9\linewidth]{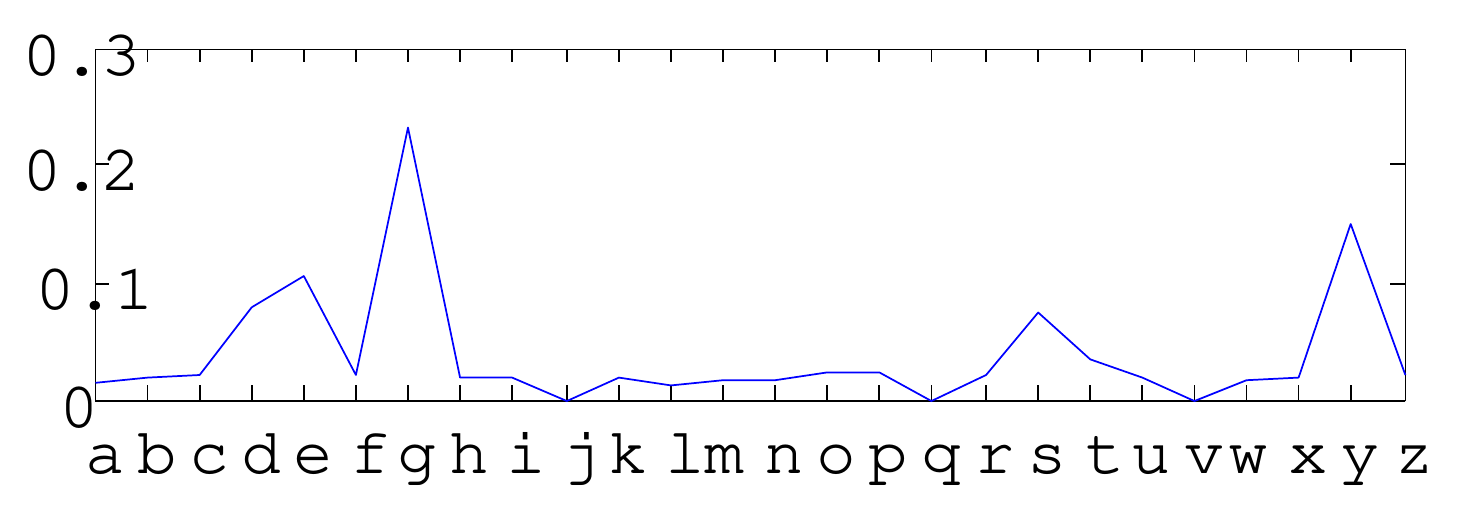}		
	\caption{Prior probability of observing a given letter as the last of a word.}
	\label{fig:lastprobs}
\end{figure}

Since words have particular grammatical rules, we exploited the fact that the last letter in a word has a much higher probability of being a consonant like \emph{g} rather than a \emph{p} for instance. So, we modeled three \acp{HMM}. One does not exploit this property and considers the entire word as a single sequence. The second models an additional last state representing these probabilities, i.e, the probability of each letter being the last in the word, thus its transition matrix is 27 by 27. These last state probabilities were computed in the same fashion as the prior probabilities, by counting and normalizing the number of times each letter appears at the end of a word. Figure~\ref{fig:lastprobs} represents this distribution. The last considers that the transition to the last word is another sequence altogether, and thus chooses the best path up until the penultimate letter, and then the most likely transition to the last one. For this model, two 26 by 26 transition matrices are used.

One important difference between other works is the way we calculate the emission probabilities which are the probabilities of a determined observation knowing the hidden state. To capture all the information that the data could give us, we opted to use each sample as a different observation, thus using the uncertainty of the classifier in a specific sample as emission probabilities. Although this approach should give better results than, for example, \cite{Velagapudi}, it requires more memory to hold all the probabilities for a given sample. Computational speed should be the same because the \ac{HMM}-Viterbi algorithm performs an indexed lookup of the emission probabilities.

\section{Base Classifiers}\label{sec:classifiers}

The base classifiers utilized were as implemented by MatLab toolboxes, save the
Parzen Window classifier, which was customly written in \emph{C}.

\subsection{Neural Networks}\label{networks}

Multi-class classification with neural networks can be implemented with either a single or multiple networks. Since the model utilized can either be \ac{OAA} or \ac{OAO}, there are several implementation approaches~\cite{murphey}. \ac{OAA} with a single neural network requires a network with as many outputs as classes, in this case 26. Utilizing multiple networks, a total of 26 networks of one output are needed. \ac{OAO} on the other hand requires training each paired class combination. That is, asumming \emph{K} classes, a total of \emph{K(K - 1)/2} networks. For our character classification problem this would correspond to 325 binary output networks. So, for ease of implementation we followed a \ac{OAA} approach using 26 binary networks. We trained 6 sets of these networks, utilizing diferent input features. The networks have a single hidden layer and the number of nodes on that layer is adjusted to the number of input parameters. The networks were trained using a batch method, Mean Square Error was used as the performance measure and scaled conjugate gradient backpropagation was the learning function.

When fed the test set, each binary network computes the probability of each particular sample belonging to the class that network models. To construct the emission matrix, these results were concatenated as rows, totaling 26 rows by a number of columns equaling the number of test samples. We simply normalized the matrix row-wise, to represent the probabilities of observing each set of features for each class.

\subsection{k-Nearest Neighbours}

The \ac{kNN} classifier was used with all the considered feature combinations. This type of classifier measures a given type of distance (euclidean, city-block, etc.) between the test and all train feature vectors. The nearest train points are selected and a decision is made based on the mode or mean of the class of the train points. In our case mode was used, as classes are discrete. 

Since that for some feature spaces some features will have a high variability in value (orientation, 0º to 180º), and others very little (pixels, 0 or 1), the contribution given to measure of distance by the features with low variance might be negligible. The optimal parameter for \emph{k}, the number of neighbors, was found by iterating for different values of \emph{k}, using the training set as known points and the validation set as test points. We imposed a maximum of 18 neighbors and decreased the value as long as there was a decrease in error. Once an increase was observed over 6 times, training stopped and the optimal value of \emph{k} was determined. We observed that error converged smoothly in function of the number of neighbours, but not completely monotonically. Euclidean distance was the used distance metric.

Calculating the emission probabilities for this case was done by counting the number of times each letter occurred in the selected set of nearest neighbors per validation sample, which composed a column of the emission matrix. The matrix was then normalized row-wise to represent probabilities of observations given classes.

%In the special case where 1 nearest neighbour is the optimal value, the emission matrix is reduced to zeros and ones. This means that

%  - prasanna only used 1 NN claiming simplicity of comparsion. what comparison?

\subsection{Naive Bayes}

\ac{NB} classifiers assume that values observed for one input variable have no correlation to any other variables. Despite the fact that this is untrue for many cases, and intuitively true in the case of handwriting recognition, \ac{NB} is known to result in acceptable classification accuracy. We tested \ac{NB} classifiers with a kernel smoothing density estimate as the underlying fitting model for each $P(C|X)$, which is the probability of observing a class marginalized to a single input feature. 

% So, we tested two variants of \ac{NB}, each one utilizing a different underlying fitting model to determine feature probability distributions. One variant used a \ac{MVMN} distribution, as it was fed with the input pixels, which is discrete data and so is appropriate for this model. The second \ac{NB} variant utilized a kernel smoothing density estimate, and so was fed with all the feature combinations as well, as it can model continuous data.

\subsection{Parzen Window}

% from the the slides:
% Kernel Regression with bandwidth h
% Small bandwidth => Higher complexity

% Como ficou o nosso? why?

% - written in C

% from the the slides:
% Kernel Regression with bandwidth h
% Small bandwidth => Higher complexity
The \ac{PW} method tries to know what is the probability distribution of the feature space directly, without assuming a parametric model. This method weighs the train samples that are nearer to the test point for which the probability is to be determined, with a larger value according to a probability distribution. This probability distribution effectively models a window that lets account for influence of some samples not others (hence the name). The most usual windows are normal distributions with a scaled identity covariance matrix. The calculation of the probability of a letter knowing the sample features $P(C|X)$ is made with the weighted sum of the samples of class $C$. The classification is then made by selecting the class that has the greater $P(C|X)$. To calculate the emission probabilities to use in the \ac{HMM}, one has just to calculate the probability of a sample knowing a class $P(X|C)$ which can be calculated using Bayesian theory on $P(C|X)$.

The big disadvantage of this method is that it is very computationally intense and so it cannot be used online. That is so much the case that the time needed to classify all our dataset with the 128 pixels didn't allow for practical training times via MatLab. That led us to develop the method in \emph{C} language to obtain and compare data.

The only parameter that affects the classification is the window bandwidth. If too large, bandwidth will make the method decide only using the prior knowledge of the data and, if to small, the resulting model will be very complex and possibly overfit to the training data. To choose the best bandwidth, we started with a large bandwidth and classified a validation set with the training set. The bandwidth could then be decreased and if the overall error in the validation set started to rise, the model was overfitting the data and in that turning point lied the best bandwidth. We have observed that the error decreased smoothly with the decrease of the bandwidth, until the overfit occurred, and that allowed us to search for the better bandwidth by halving the bandwidth search space in each iteration making for a much faster convergence. This bandwidth, or H parameter, is the length of the sides of a hypercube in the same dimensional space as the feature space. So, if the feature space is in $\mathbb{R}^3$ the bandwidth will be the equivalent of the volume in that space, which is $H^3$ or in general $H^d$.

\section{Classification Performances}\label{sec:results}

% tempos ao correr KNN no meu PC mais lentos do que correr no apps?? why.
% why is main different from recorded runs?

Table~\ref{tab:resultstab} displays the train an test classification accuracies for the used base classifiers for each particular feature set, followed by the test accuracy achieved by error correction. \emph{HMM(1)} denotes the accuracy with a \ac{HMM} modeling the entire word as a sequence, \emph{HMM(2)} is the accuracy attained by modeling the last word probabilities as an additional state and \emph{HMM(3)} the accuracy for the \ac{HMM} modeling the last transition as a separate sequence. The \emph{h/k/Ep.} column applies to the \ac{PW}, \ac{kNN} and \ac{NN} classifiers, and represents the optimal value for \emph{h}, \emph{k} and the average number of training epochs respectively. Lastly, the training and test times in minutes are presented.

\begin{table*}
\centering
\caption{Classification results for several features and classifier combinations. For \acp{NN}, the parentheses indicate number of nodes. The feature sets are as follows: \emph{a)} contains the only the centroid and eccentricity data; \emph{b)} adds to this the number of holes and objects in the letter; \emph{c)} adds the pixel count per quadrant; \emph{d)} adds the perimeter, length of major and minor ellipse axes and ellipse orientations; \emph{e)} adds the 128 preprocessed pixels; \emph{f)} contains only the original unprocessed data set; \emph{g)} contains the preprocessed dataset; \emph{h)} adds to that with the respective extracted orientation information}
\begin{tabular}[t]{ c | c || c | c | c | c | c | c | p{1cm} | p{1cm} }
  \multicolumn{4}{c}{} & \multicolumn{3}{|c|}{HMM Model} & \multicolumn{2}{c}{}\\
  \bf{Classifier} & \bf{Features} & \bf{Train Acc.} & \bf{Test Acc.} & \bf{HMM(1)} & \bf{HMM(2)} & \bf{HMM(3)} & \bf{h/k/Ep.} & \bf{Train Time} & \bf{Test Time}\\
  \hline
  K-NN & a) & 42.7 & 42.1 & 50.8 & 52.1 & 51.7 & 17 & 0.09 & 0.03\\
  K-NN & b) & 53.4 & 53.1 & 63.1 & 63.9 & 63.7 & 17 & 0.16 & 0.03\\
  K-NN & c) & 64.1 & 63.7 & 72.6 & 73.3 & 73.4 & 11 & 0.34 & 0.04\\
  K-NN & d) & 72.0 & 70.8 & 79.8 & 80.3 & 80.5 & 9 & 2.30 & 0.15\\
  K-NN & e) & 77.9 & 77.3 & 79.0 & 79.2 & 79.0 & 1 & 19.13 & 1.04\\
  K-NN & f) & 78.5 & 78.3 & 79.6 & 80.0 & 80.2 & 1 & 15.83 & 0.87\\
  K-NN & g) & 74.7 & 74.5 & 82.7 & 83.0 & 83.7 & 5 & 15.33 & 0.83\\
  K-NN & h) & 69.9 & 69.3 & 70.5 & 71.0 & 71.2 & 1 & 16.77 & 0.93\\
  \hline
  \bf{Average} & & 66.7 & 66.1 & 72.3 & 72.9 & 73.0 & 7.8 & 8.7 & 0.49\\
  \hline
  \hline
  NN (12) & a) & 43.7 & 44.1 & 53.3 & 54.2 & 53.6 & 56.31 & 27.9 & 0.02\\
  NN (16) & b) & 47.0 & 47.5 & 53.9 & 55.3 & 55.4 & 71.42 & 36.1 & 0.02\\
  NN (28) & c) & 69.1 & 68.8 & 78.2 & 78.7 & 78.3 & 84.54 & 48.4 & 0.03\\
  NN (35) & d) & 78.9 & 70.9 & 73.6 & 74.3 & 72.3 & 58.50 & 42.9 & 0.07\\
  NN (35) & e) & 85.7 & 79.3 & 84.0 & 84.5 & 84.8 & 52.58 & 37.5 & 0.06\\
  NN (64) & f) & 85.5 & 78.4 & 82.8 & 83.4 & 82.6 & 53.92 & 112.04 & 0.08\\
  \hline
  \bf{Average} & & 68.32 & 64.83 & 70.97 & 71.73 & 71.17 & 62.86 & 50.81 & 0.05\\
  \hline
  \hline
  PW & a) & 42.9 & 43.3 & 52.4 & 53.6 & 52.0 & 0.1345 & 0.44 & 0.35\\
  PW & b) & 50.4 & 51.6 & 61.8 & 62.4 & 61.0 & 0.4358 & 0.32 & 0.31\\
  PW & c) & 63.0 & 63.1 & 72.1 & 72.5 & 71.8 & 0.9143 & 0.30 & 0.32\\
  PW & d) & 70.8 & 72.3 & 81.1 & 81.7 & 81.9 & 1.0965 & 0.63 & 0.65\\
  PW & e) & 78.9 & 78.7 & 85.8 & 86.1 & 86.3 & 1.0094 & 1.05 & 1.45\\
  PW & f) & 79.0 & 79.6 & 89.4 & 89.8 & 89.9 & 0.9967 & 0.59 & 1.07\\
  PW & g) & 76.6 & 75.3 & 84.6 & 84.8 & 85.3 & 0.9948 & 0.64 & 1.07\\
  PW & h) & 69.5 & 70.2 & 79.7 & 80.1 & 79.3 & 1.0002 & 1.39 & 1.46\\
  \hline
  \bf{Average} & & 66.39  & 66.76 & 75.86 & 76.38 & 75.94 & 0.82 & 0.67 & 0.84 \\
  \hline 
  \hline
  % NB(mvmn) & f) & 6 & 4 & 5 & 6 & 1 & 1 & 1 & 1\\
  % NB(mvmn) & g) & 6 & 4 & 5 & 6 & 1 & 1 & 1 & 1\\
  % \hline  
  % NB(ker.) & a) & 45.5 & 45.3 & 55.5 & 56.3 & 55.6 & 1 & 0.55 & 1.07\\
  % NB(ker.) & b) & 47.9 & 48.2 & 55.8 & 57.8 & 59.0 & 1 & 1.12 & 2.23\\
  % NB(ker.) & c) & 57.4 & 56.9 & 67.8 & 69.5 & 70.6 & 1 & 1.89 & 3.76\\
  % NB(ker.) & d) & 62.7 & 62.0 & 75.2 & 76.0 & 77.2 & 1 & 2.60 & 5.17\\
  % NB(ker.) & e) & 44.6 & 44.3 & 46.3 & 46.4 & 48.5 & 1 & 41.10 & 78.11\\
  % NB(ker.) & f) & 41.6 & 41.4 & 42.2 & 42.3 & 43.3 & 1 & 35.07 & 66.63\\
  % NB(ker.) & g) & 34.4 & 34.2 & 34.3 & 34.5 & 34.7 & 1 & 37.25 & 78.21\\
  % NB(ker.) & h) & 6 & 4 & 5 & 6 & 1 & 1 & 1 & 1\\
  NB & a) & 45.5 & 45.3 & 55.5 & 56.3 & 55.6 & - & 0.55 & 1.07\\
  NB & b) & 47.9 & 48.2 & 55.8 & 57.8 & 59.0 & - & 1.12 & 2.23\\
  NB & c) & 57.4 & 56.9 & 67.8 & 69.5 & 70.6 & - & 1.89 & 3.76\\
  NB & d) & 62.9 & 62.0 & 75.2 & 76.0 & 77.2 & - & 2.60 & 5.17\\
  NB & e) & 44.6 & 44.3 & 46.3 & 46.4 & 48.5 & - & 41.10 & 78.11\\
  NB & f) & 41.6 & 41.4 & 42.2 & 42.3 & 43.3 & - & 35.07 & 66.63\\
  NB & g) & 34.4 & 34.2 & 34.3 & 34.5 & 34.7 & - & 37.25 & 78.21\\
  NB & h) & 34.8 & 34.5 & 34.8 & 35.0 & 35.2 & - & 41.10 & 78.22\\  
  \hline
  \bf{Average}& & 46.11 & 45.9 & 51.5 & 52.2  &  53.0  & -  & 20.1  & 39.2\\
  \hline
  \hline
  \bf{Total Average}&& 61.9 & 60.9 & 67.6 & 68.3 & 68.3 & 35.33 & 20.07 & 10.1\\
\end{tabular}
% a) feats_6
% b) feats_8
% c) feats_12
% d) feats_13
% e) feats_13_and_pixels
% f) pixels_128
% g) pixels_128_rotated
% h) pixels_128_rotated_w_ori
\label{tab:resultstab}
\end{table*}

%According to what we have observed, preprocessing the pixelmaps by rotation do not result in improvementsl in any of the used classifiers, degrading performance by an average of 8.4\%. This was somewhat expected because information was being removed from the data. However, performance gain was expected by supplying rotation information as a separate feature, as this would orthogonalize the feature space and potentialy help the classification process. Again, the orthogonalization did not help and even decreased the performance for every case. Rotating the pixels did not perform better possibly because the loss of data from the rotation was bigger than the gain from extracting the orientation. Separately fed orientation information may have in fact made the distinction more ambiguous because different letters share orientations.

According to the results, preprocessing the pixelmaps by rotation did not cause improvements in any of the used classifiers, degrading performance by an average of 8.4\%. This was somewhat expected because information was being removed from the data. However, performance gain was expected by supplying rotation information as a separate feature, as this would orthogonalize the feature space and potentially help the classification process. However, this was not the case possibly because the loss of data from the rotation was bigger than the information gain from extracting the orientation. Separately fed orientation information may have in fact made the distinction more ambiguous because different letters share orientations. 

% In fact, After applying viterbi error correction, performance with feature set \emph{g)} only improves for \acp{kNN}, regardless of the type of the \ac{HMM} model. Note that the number of neighbours is 5 as opposed to 1 for feature set \emph{f)}. Feature set \emph{h)} which contains orientation as a separate input does not improve recognition either, decreasing the rates for every case, \verificar{following that happens with the rest of the classifiers.}

The best base classifier performance of 79.6\% was found for a \ac{PW} classifier with feature set \emph{f)}. A close second was 79.3\% for a \ac{NN} with 35 nodes using feature set \emph{e)}. The worst case was observed for \ac{NB}, with 34.2\%.

The classifier that benefited the most from the \emph{HMM(2)} model was the \ac{NB} classifier with recognition rate increase of 24\% for the feature set \emph{a)}. The smallest recognition increase was also for the \ac{NB} classifier with a rate increase of 0.88\% for the feature set \emph{g)}. In average, the classifier that benefited the most from this \ac{HMM} model was the \ac{PW} and the one that benefited the least was the \ac{NN}, with increases of 14\% and 10\% respectively. The increase with this \ac{HMM} model, which includes the last transition state, is consistently higher than the performance gains from \emph{HMM(1)}. For this model, these values were 13.48\% and 9.37\%, respectively, and for the same classifiers.

A recognition rate decrease is never verified by utilizing \acp{HMM}. However, performance seems to consistently improve less with the increase of the classifier model complexity. For the smallest feature space (i.e.,  lowest complexity) the average increase amongst classifiers for \emph{HMM(2)} results average improvement of 23,7\%, which corresponds to the best increase for all classifiers. The smallest increase occurs for features spaces that contain the pixelmap data, but there is not a single discriminate feature space for which increase is lowest for all classifiers. This may be easy to explain if we consider that the more complex models reflect more certainty in the emission probabilities, thus resulting in less misclassification errors to correct.

Quick additional tests we performed with \acp{NN} showed that a data division ratios of 70, 15 and 15\% for the train, validation and test set resulted in better classification performance, at a peak of 92,35\%, for a \ac{NN} of 35 nodes and feature set \emph{f)}. We may extrapolate that using a larger training set might have resulted in performance increases for all classifiers. However, we utilized out data division of 1/3 for each set to ensure that validation and test sets were not too reduced.
 
 % \verificar{- optimal is 1 neighbour for 128 pixels because any one change in any pixel results in a nearest neighbour. This means that there is a wide selection space for the nearest neighbor. This is equivalent to choosing the pixelmap that is the most similar as a near neighbour (i.e. with the least change in pixels). Effectively , this is equivalent to subtracting the pixelmaps, and choosing the class for which the difference is smallest (in absolute value).
 % ...this is the same as overfitting the classifier considerably}
 
%%%%%%%%%%%%%%%%%
\section{Conclusions}\label{sec:conclusions}

This study presented results of base classifier recognition rates for a handwritten dataset consisting of 26 characters. Four different classifiers were used with eight different feature sets extracted from the dataset. The classifier decisions were then corrected via \ac{HMM} models that explored neighboring relations between words. In every case, \acp{HMM} models improved classification. Modeling the last transition of a word further increased the performance also for every case.

%Classification aided by Viterbi error correction improved performance in every tested case. 
Curiously, the \emph{HMM(3)} model has better results over the \emph{HMM(2)} model for about half of the cases, which is not consistent with the mathematical model of \acp{HMM}. This may be due to the small size of the dataset, but we cannot state this with certainty.

%- should our input space be bigger? by how much and containing what?
Our input space was comparatively small in relation to other OCR approaches, but a reasonable classification performance is still achieved. Still, a more thorough feature extraction method could have aided detection~\cite{4798415}. We observe that feature spaces containing the pixelmaps as inputs result in better performance, even when utilizing one hold-out validation, versus 10 cross-fold, as was done in~\cite{Velagapudi}.

%- could we even have gotten more features from the reduced input set that we had?

\bibliographystyle{ieee}
\bibliography{OCR}

\end{document}